\documentclass[runningheads]{llncs}
\usepackage{graphicx}
\usepackage{amsmath,amssymb} % define this before the line numbering.
\usepackage{color}
\usepackage{subfigure}
\usepackage{multirow}
\usepackage[width=122mm,left=12mm,paperwidth=146mm,height=193mm,top=12mm,paperheight=217mm]{geometry}
\begin{document}
% \linenumbers
\pagestyle{headings}
\mainmatter

\title{Auto-Context R-CNN} % for Accurate and Robust Object Detection
% or: Auto-Context R-CNN

\titlerunning{Auto-Context R-CNN}
\authorrunning{B. Li et al}

\author{Bo Li$^\dagger$, Tianfu Wu$^\ddagger$\thanks{T. Wu is the corresponding author}, Lun Zhang$^\dagger$ and Rufeng Chu$^\dagger$}

\institute{$^\dagger$YunOS BU, Alibaba Group\\
$^\ddagger$ Department of Electrical and Computer Engineering and the Visual Narrative Cluster,
North Carolina State University\\
\email{ \{shize.lb,lun.zhangl,rufeng.churf\}@alibaba-inc.com, tianfu\_wu@ncsu.edu }
}

\maketitle

\begin{abstract}
Region-based convolutional neural networks (R-CNN)~\cite{fast_rcnn,faster_rcnn,mask_rcnn} have largely dominated object detection. Operators defined on RoIs (Region of Interests) play an important role in R-CNNs such as RoIPooling~\cite{fast_rcnn} and RoIAlign~\cite{mask_rcnn}. They all only utilize information inside RoIs for RoI prediction, even with their recent deformable extensions~\cite{deformable_cnn}. 
Although surrounding context is well-known for its importance in object detection, it has yet been integrated in R-CNNs in a flexible and effective way. Inspired by the auto-context work~\cite{auto_context} and the multi-class object layout work~\cite{nms_context}, this paper presents a generic context-mining RoI operator (i.e., \textit{RoICtxMining}) seamlessly integrated in R-CNNs, and the resulting object detection system is termed \textbf{Auto-Context R-CNN} which is trained end-to-end. The proposed RoICtxMining operator is a simple yet effective two-layer extension of the RoIPooling or RoIAlign operator. Centered at an object-RoI, it creates a $3\times 3$ layout to mine contextual information adaptively in the  $8$ surrounding context regions on-the-fly. Within each of the $8$ context regions, a context-RoI is mined in term of discriminative power and its RoIPooling / RoIAlign features are concatenated with the object-RoI for final  prediction. \textit{The proposed Auto-Context R-CNN is robust to occlusion and small objects, and shows promising vulnerability for adversarial attacks without being adversarially-trained.} In experiments, it is evaluated using RoIPooling as the backbone and shows competitive results on Pascal VOC, Microsoft COCO, and KITTI datasets (including $6.9\%$ mAP improvements over the R-FCN~\cite{rfcn} method on COCO \textit{test-dev} dataset and the first place on both KITTI pedestrian and cyclist detection as of this submission). % can not mention first place?          
\end{abstract}

% TODO: 
% 1) Figures: 
% 2) Justification: context layout-->Zhuowen's auto-context and Ramaman's multi-object context (discuss Jifeng's deformable and SPN)

\section{Introduction}
\subsection{Motivations and Objectives}
State-of-the-art deep learning based object detection systems have been largely dominated by the region-based convolutional neural networks (R-CNN)~\cite{fast_rcnn,faster_rcnn,mask_rcnn}. R-CNNs consist of two components: (i) A class-agnostic region proposal component (i.e., objectness detection) is used to reduce the number of candidates to be classified since the sliding window technique is practically prohibitive. The proposals, called RoIs (region-of-interest) are generated  either by utilizing off-the-shelf objectness detectors such as the selective search~\cite{SS}, Edge Boxes~\cite{edge_boxes} and BING~\cite{BING} or by learning an integrated region proposal network (RPN)~\cite{faster_rcnn} end-to-end. (ii) A prediction component (i.e., head classifier) is used to classify all the RoIs and to regress bounding boxes for accurate detection. The two components can be either separately or end-to-end trained. To accommodate different shapes of RoIs, operators such as RoIPooling~\cite{fast_rcnn} and RoIAlign~\cite{mask_rcnn} need to be adopted to compute equally-sized feature maps for the prediction component. The RoI operators first divide a RoI of any shape into a predefined grid of $ph\times pw$ cells (e.g., $ph=pw=7$ or $14$ typically used in practice). Each cell in the grid is described by a vector of the same dimensionality as the feature map through spatial max/average-pooling inside the cell. Then, the concatenated vector is used as input to the head classifier (such as a fully-connected layer).  With the recent deformable extensions~\cite{deformable_cnn}, although each cell has the potential to be placed outside the RoI through the feedforward computed displacement, the experimental results~\cite{deformable_cnn} showed that the cells are mostly deformed toward inside the RoI. \textit{Exploiting information mainly inside an object RoI not only may not fully explore the power of R-CNN based object detection systems (e.g., handling small objects and occlusion), but also may be less vulnerable to adversarial attacks targeted object regions~\cite{attackRCNN,AttackDetectors}}. 

It is well known that contextual information is important in object detection and visual  recognition~\cite{ctxPriming,ctxEmpirial,ctxTinyImages,ThreeChannel-IJCV}. To integrate surrounding contextual information in R-CNNs, different methods have been investigated: global context (i.e., RoIPooling using the whole image lattice as a RoI)~\cite{resNet}, inside-outside networks utilizing spatial recurrent neural network for gathering contextual information~\cite{ion}, and local context pooled from predefined context regions w.r.t. an object RoI~\cite{mrcnn}. We are interested in methods of mining surrounding contextual information explicitly and on-the-fly in learning and inference  which can be seamlessly integrated in R-CNNs.  \textit{Inspired by the auto-context work~\cite{auto_context} and the multi-class object layout work~\cite{nms_context}, we propose a simple yet effective  extension to the RoIPooling or RoIAlign operators, which discriminatively mines surrounding contextual information for improving object detection performance.} The resulting object detection system is termed \textbf{Auto-Context R-CNN} which is trained end-to-end. As a by-product, it also shows promising vulnerability for adversarial attacks without being adversarially-trained. 

\subsection{Method Overview}
Our Auto-Context R-CNN extends Faster R-CNN~\cite{faster_rcnn} with the RoIPooling~\cite{fast_rcnn} or RoIAlign~\cite{mask_rcnn} operator substituted by the proposed RoICtxMining operator. 

\setlength{\textfloatsep}{4pt}
\begin{figure}[t]
	\centering
	\begin{minipage}[t]{0.53\textwidth}
		\centering
		\includegraphics[width=\textwidth]{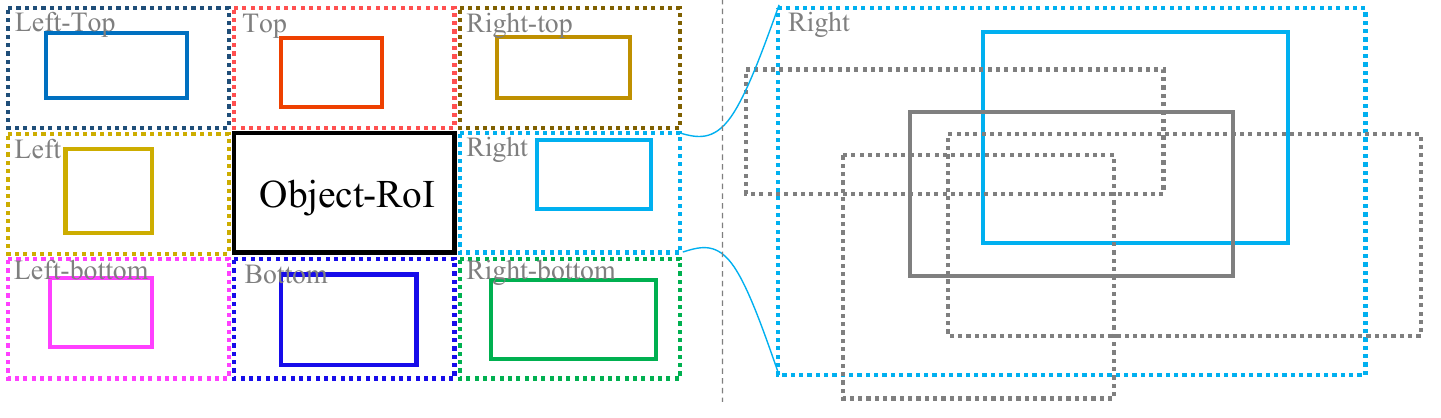}
	\end{minipage}
    %\hspace{3cm}
	\begin{minipage}[t]{0.44\textwidth}
		\centering
		\includegraphics[width=\textwidth]{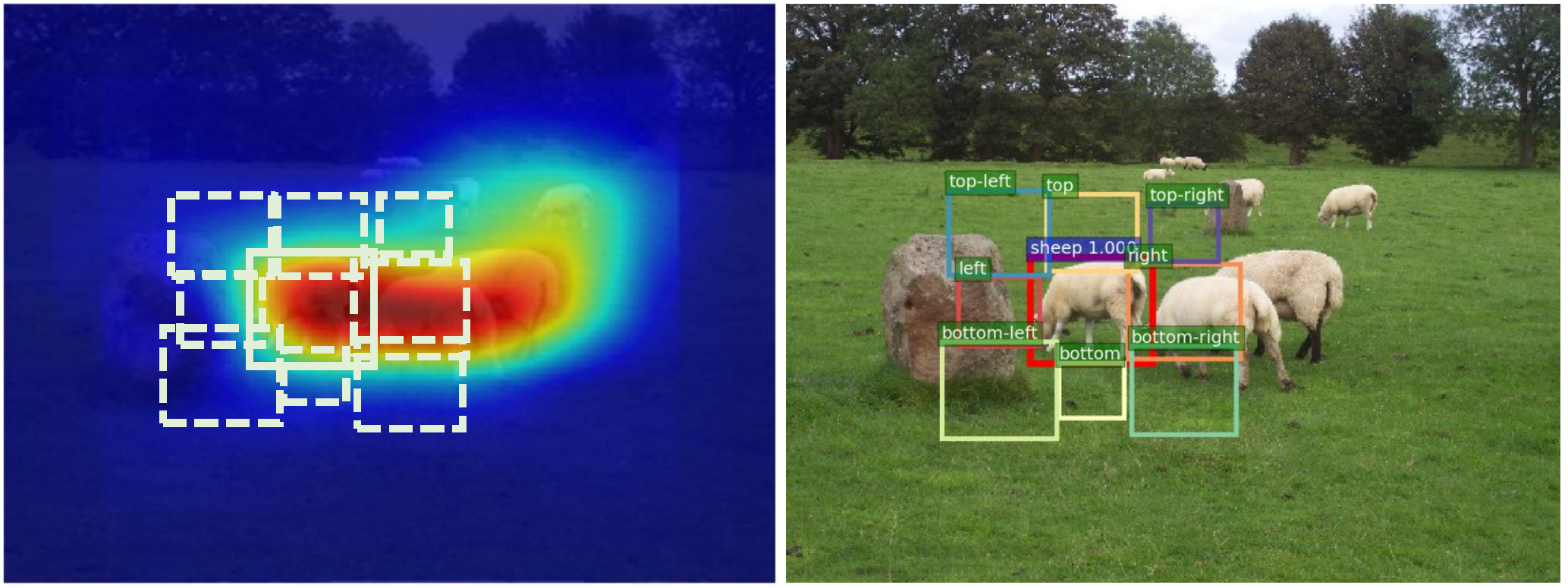}
		%\caption{An example of context mined by the RoICtxMining operator.} \label{fig:RoICtxMiningDemo}
	\end{minipage}
    \caption{Illustration of our RoICtxMining operator (left) and an example in detection (right). Centered at an object-RoI, $8$ surrounding context-RoIs are minded in corresponding regions of a $3\times 3$ layout. See text for detail.} \label{fig:RoICtxMining}
    %\vspace{-10pt}
\end{figure}

%Object detection via Faster R-CNN in an image consists of two stages: generating category-agnostic RoIs (i.e., bounding box proposals, so-called objectness detection) and classifying each RoI into one of the $C+1$ categories (e.g., the $C=20$-class in the PASCAL VOC dataset~\cite{pascal} and a background class) in terms of prediction scores with the RoI regressed for more accurate detection. The former is implemented by the RPN sub-network and the latter is done through the RoI-wise prediction sub-network, both sharing a feature backbone. They are trained end-to-end. 

Fig.~\ref{fig:RoICtxMining} illustrates the proposed RoICtxMining operator which is a two-layer extension of the traditional RoI operator. An object proposal is represented by an object-RoI and $8$ context-RoIs which capture contextual information in the $8$ spatial neighbors (left, right, top, bottom, left-top, right-top, left-bottom and right-bottom). Centered at the object-RoI, we define a $3\times 3$ grid with each cell being the same shape as the object-RoI, which defines the contextual layout similar in spirit to the auto-context work~\cite{auto_context}, the multi-class object layout work~\cite{nms_context} and the recent deep context in unsupervised representation learning~\cite{DeepContext}. Each of the $8$ context-RoIs is mined w.r.t. the corresponding cell. Feature representations of the object-RoI and the context-RoIs are computed by either RoIPooling or RoIAlign operator, then concatenated together for final prediction.   

To mine the context-RoI in a surrounding neighbor cell (e.g., the cell to the right of the object RoI in Fig.~\ref{fig:RoICtxMining}), we first define the pool of context-RoI proposals. We have some high-level assumptions: context-RoI proposals are largely confined inside the cell and not too small, and only one context-RoI proposal in the pool will be mined. In defining the pool, we use the center RoI as the anchor RoI (see the solid gray RoI in the enlarged cell in Fig.~\ref{fig:RoICtxMining}). The anchor RoI is of half width and half height of the cell. The pool includes all RoIs whose intersection over union (IoU) with the anchor RoI are greater than or equal to a threshold ($0.3$ used in our experiments, see some examples illustrated by the dotted gray RoIs in the enlarged cell in Fig.~\ref{fig:RoICtxMining}). Each context-RoI proposal is described by a feature vector computed by RoIPooling or RoIAlign operator. To mine the most discriminative context-RoI in the pool, we introduce a class-agnostic sub-network (e.g., a simple $1\times 1$ convolution or equivalently a fully connected layer with one output) which computes the score (goodness of contextual information) for each context-RoI proposal and select the one with the maximum score as the mined context-RoI.   

The proposed method of mining context-RoI in a surrounding neighbor cell is related to the latent part placement approach in the deformable part-based models (DPMs)~\cite{DPM}. The mined context-RoIs can be treated as latent contextual parts. The proposed method is also related to the recent deformable convolution method~\cite{deformable_cnn} which computes the displacement in a bottom-up way without explicit control. Here, the pool of context-RoI proposals is defined with explicit control of the range in a top-down manner w.r.t. a given object-RoI and the proposals cover not only displacement but also shape changes.  

In experiments, we test our Auto-Context R-CNN using RoIPooling as the backbone RoI operator. It shows competitive results on Pascal VOC 2007 and 2012~\cite{pascal}, Microsoft COCO~\cite{coco}, and KITTI datasets~\cite{kitti}. We compare with a series of state-of-the-art deep learning based object detection systems (including recent Faster R-CNN variant, R-FCN~\cite{rfcn} with $6.9\%$ mAP improvement on COCO \textit{test-dev} dataset). Our model also won the first place on both KITTI pedestrian and cyclist detection (anonymous submission). Our model also shows promising vulnerability robustness to object adversarial attacks. 

\section{Related Work and Our Contributions}
We first briefly review R-CNN based detection systems and context models, then introduce our contributions to the computer vision community.

\textbf{Region-based Detection Models.}
Recently, region-based detection models \cite{rcnn,rfcn,faster_rcnn,deformable_cnn,mask_rcnn} have dominated the object detection field, as they obtain leading accuracies on popular benchmarks \cite{pascal,imagenet,coco}. 

Starting from the seminal work - R-CNN \cite{rcnn}, which consists of an object proposal module and region-based CNN classifier. R-CNN was extended by \cite{sppnet,fast_rcnn} to allow extracting RoI specific features using RoIPooling, increasing both the detection speed and accuracy. Faster R-CNN  \cite{faster_rcnn} replaced the separating object proposal module (e.g., Selective Search \cite{SS}) with a region proposal network (RPN), leading to a significant speedup for proposal generation. R-FCN \cite{rfcn} further improved the speed using position-sensitive RoIPooling, making the whole network  fully convolutional. Recently, He et al. \cite{mask_rcnn} introduced RoIAlign that properly aligning extracted features with positions of input region proposals, leading to further improvement on detection accuracy.
Besides, there are several follow-up works \cite{ohem,MS-CNN,FPN,megdet} making the region-based detection models more robust and flexible. 
%However, the main stream of region based detectors focuses on the object RoIs, while contexts outside the RoIs have attracted little attentions.

\textbf{Context Modeling.}
The role of context has been well exploited in recognition and detection \cite{auto_context,nms_context,carAOG,guangchen_cvpr,hoiem06,torralba,laptev15}. Specifically, \cite{auto_context} utilized the classification confidence as context information, and learn the context model in an iterative way, showing significant improvements over Conditional Random Fields (CRFs) and Belief Propagation (BP) on context modeling.  Desai et al. \cite{nms_context} proposed to model the object-object semantic relationship as context in a $3 \times 3$ cells, leading to improved performance over deformable part-based models (DPM) \cite{DPM} on PASCAL VOC dataset. 

For generic object detection with CNN, numerous works have been proposed.
Sermanet et al. \cite{sermanet} utilized two contextual regions centered on each object for pedestrian detection. In \cite{scalable}, features of both specific regions and the entire image are used to improve region classification. For ResNets-based Faster R-CNN model \cite{resNet}, only the entire image contextual feature was utilized.  Meanwhile, \cite{deepid} utilized classification scores of the whole image to aid object recognition.
\cite{mrcnn} utilized ten contextual regions around each object with different crops to improve the localization of objects, while \cite{multipath,gbdnet} utilized four contextual regions.
\cite{ion} used spatial RNNs to compute contextual features in object detection.
Though those methods provide a simple way to integrate contextual information, the context is manually predefined. Besides, the improvement is usually modest or marginal.

Recently, \cite{non_local} use the non-local operations to capture long-range dependencies of any two pixels in an image, which implicitly captures the pixel-to-pixel relationship across the entire image domain, it is complementary to local contexts for higher-level semantic object detection.
\cite{auto_net} iteratively utilized the posterior distribution of labels along with image
features, similar to \cite{auto_context}, for brain extraction in magnetic resonance imaging.
Recurrent rolling convolution network \cite{RRC} aggregates features from both top-down and bottom-up layers, though is ``deep in context", the training process is slow and hard to converge.

%Inspired by \cite{auto_context,nms_context}, we provide an efficient and effective way to automatically mining contextual information in the $3 \times 3$ cells centered at each object. Different from previous works that manually defining context regions, we dynamically learn the discriminative context around each object. In experiments, we find the learned context model gives significant improvements on PASCAL VOC, Microsoft COCO and KITTI benchmarks. 

\iffalse{
\textbf{Accurate Localization.}
Many works resort to auxiliary features to aid object localization. \cite{fidler} incorporated segmentation cues with DPM, \cite{hoiem_loc} utilized color and edge features, \cite{Schulter_2014_CVPR} used the height prior of an object. To improve the localization ability of R-CNN,
\cite{yuting} used Bayesian optimization to refine the bounding box proposals and trained the CNNs with a structured loss. \cite{locNet} assigned probabilities on each row and column of a search region to get accurate object positions. Those works are complementary with our framework.
In this paper, we propose an extra training step for high precise object localization. In the literature, [iterative rpn] provides a cascade and iterative method to get more accurate RPN proposals, though similar to our methods, we utilize RoI proposals generate by the region classification module rather than RPN, and our method is much more simple yet effective.
}\fi

This paper makes the following three main contributions to object detection. 

%\begin{itemize} [leftmargin=*]
%\item
(i) We propose a generic context-mining RoI operator (i.e., RoICtxMining) seamlessly integrated in R-CNN based object systems;

%\vspace{-2mm}
(ii) Instead of manually assigning contexts around objects, we present a novel method that automatically mining context for object detection;

(iii) Our model obtains significant improvement over state-of-the-art comparable R-CNN based detection systems on PASCAL VOC 2007 and 2012, Microsoft COCO, and KITTI pedestrian and cyclist datasets (first place in the leadboards as of submission).
%\vspace{-2mm}
%\end{itemize}
%\vspace{-2mm}

\begin{figure} [t]
\centering
%\framebox
{\includegraphics[width = 0.75\textwidth]{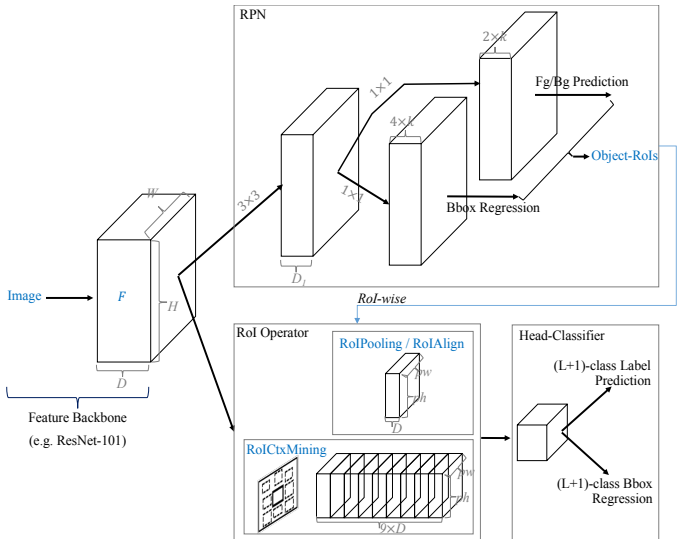}}
\caption{Network architecture of the proposed Auto-Context R-CNN built on top of the Faster R-CNN framework~\cite{faster_rcnn}. See text for details.}
\label{fig:network} %\vspace{-5mm}
\end{figure} 

\section{Auto-Context R-CNN}
In this section, we present details of the proposed Auto-Context R-CNN. To be self-contained, we first briefly introduce the Faster R-CNN~\cite{faster_rcnn}. Then, we elaborate on the proposed RoICtxMining operator and the learning method. Fig.~\ref{fig:network} illustrates the network architecture.

\subsection{Overview of Faster R-CNN}
Faster R-CNN is an end-to-end learnable object detection framework. Object detection via Faster R-CNN in an image consists of three components: the feature backbone (such as the ResNet~\cite{resNet}), the  category-agnostic RoI proposal sub-network (i.e., the RPN~\cite{faster_rcnn}), and the RoI-wise prediction sub-network (i.e., the R-CNN head classifier). The feature backbone is shared by the RPN and the R-CNN head classifier. 

\textit{The feature backbone} computes a feature map for an input image $x$. Denote by $F$ the feature map which is represented by a tensor of $D\times H \times W$ where $D$ is the dimensionality of the feature space, and $H$ and $W$ the height and width of the feature map respectively (which are usually much smaller than those of the input image $x$ due to sub-sampling in the backbone ConvNet). 

\textit{The RPN} is used for objectness detection (binary classification), that is to generate a set of class-agnostic RoIs. To that end, a small number $k$ of anchor boxes are predefined in a translation invariant way (e.g., $k=9$ anchor boxes are used covering the combination between $3$ aspect ratios and $3$ sizes). By placing the anchor boxes in the feature map, there are $k\times H\times W$ RoI proposals (which are a subset of sliding windows). Based on IoU with annotated object bounding boxes, RoI proposals are assigned as either positives or negatives. For positive RoI proposals, bounding box regression parameters are also computed for better localization. The RPN is a lightweight subnetwork. It takes the feature map $F$ as input and usually applies a $3\times 3$ convolution, followed by two branches of  $1\times 1$ convolution for binary classification and bounding box regression respectively for the $k$  anchor boxes placed at each position in the feature map. For RoIs classified as positives (i.e., object-RoIs), after bounding box regression, they go through non-maximum suppression (NMS) and the top $N_{RPN}$ object-RoIs will be kept (e.g., $N_{RPN}=2000$ is typically used).

\textit{The R-CNN head classifier} is used to classify the object-RoIs from the RPN into $L+1$ categories (e.g., $L=20$ in PASCAL VOC and a generic background class). To that end, a RoI operator is first used to compute equally-sized (denoted by $ph\times pw$, typically $ph=pw=7$) feature representation for all object-RoIs which usually have different shapes and sizes. Denote by $r$ an object-RoI of size $h\times w$ in the feature map $F$. After the RoI operator, $r$ is represented by a RoI map, denoted by $F(r)$, with dimensionality $D\times ph \times pw$ computed from the feature map $F$. The RoIPooling operator~\cite{fast_rcnn} and RoIAlign operator~\cite{mask_rcnn} are widely used. After the RoI operator, the R-CNN head classifier has two branches too, one for $L+1$-class  prediction and the other for $L+1$-class bounding box prediction. They can be implemented by a shared fully connected layer followed by another fully-connected layer for each branch. Or, they are implemented by some heavier design such as the final stage of ResNet~\cite{resNet} followed by a fully-connected layer for each branch.

\subsection{The RoICtxMining Operator}
Popular RoIPooling and RoIAlign operators utilize information only inside object-RoIs. To leverage surrounding contextual information, our RoICtxMining operator is a simple two-layer extension for them. 

As illustrated in Fig.~\ref{fig:RoICtxMining}, given an object-RoI $r$, we place a $3\times 3$ grid centered at $r$ with  all cells being the same shape as $r$. Denote by $R_c$ and $r_c$ a surrounding cell in the $3\times 3$ grid and the corresponding context-RoI mined in the cell respectively where $c\in \{left, top, right, bottom, left-top, right-top, right-bottom, left-bottom\}$. With the $8$ mined context-RoIs, an object-RoI is then described by the concatenated RoI maps of dimensionality $9\cdot D\times ph \times pw$. All the $9$ RoI maps are computed by either RoIPooling or RoIAlign. As illustrated in Fig.~\ref{fig:network}, this augmented RoI representation is the only change to the Faster R-CNN framework. 

To mine the context-RoI $r_c$ in the cell $R_c$, we adopt the idea similar in spirit to the RPN. We have two components: 
\begin{itemize} 
\item \textit{Defining the candidate pool of context-RoIs.} We first define the set of candidate context-RoIs, denoted by $\Omega = \{b_0, \cdots, b_n\}$ where $b_0$ represents the anchor RoI which is placed
at the center of the cell $R_c$ with half height and half width of the cell. Then, we enumerate all $b_i$'s which satisfy i) the short edge is not shorter than one third of the short edge of the cell and the long edge is not longer than the long edge of the cell (i.e., a context-RoI is assumed to be not too small or not too big w.r.t. the surrounding cell itself), and ii) the IoU between a $b_i$ and the anchor $b_0$ is greater than or equal to $0.3$ (i.e., context-RoIs are assumed to be placed not too far from the center of the cell). We note that the  assumptions for the size and the distance to center can be treated as hyper-parameters. We keep them fixed for simplicity in our experiments. 
\item \textit{Selecting the best context-RoI in the pool.} Each candidate context-RoI $b_i$ is represented by a RoI map using either RoIPooling or RoIAlign. To select the most discriminative one, we introduce a fully-connected layer with one output which computes the score for each candidate context-RoI, and we select as the context-RoI  $r_c$ the one with the best score (i.e., max-pooling across the entire pool for each surrounding cell individually). The fully connected layer is shared among all the $8$ surrounding cells. 
\end{itemize}

After the context-RoI $r_c$ is mined for each surround cell in the $3\times 3$ layout, we concatenate their RoI maps with that of the object-RoI, then apply the R-CNN head classifier for final prediction and bounding box regression.

\subsection{Learning Auto-Context R-CNN}
In this section, we briefly present the multi-task formulation in parameter learning, which are the same as in Faster R-CNN~\cite{faster_rcnn}.  %Then we present an effective iterative training procedure, as well as some implementation details.

Both the RPN subnetwork and the R-CNN subnetwork are trained using a multi-task loss including the classification loss and the bounding box regression loss as in \cite{fast_rcnn,faster_rcnn}. The objective function is defined as follows:
\begin{align}
\mathcal{L}(\{p_j, \ell_j\}, \{t_j, t_j^*\}) = &{1\over N_{cls}}\sum_j \mathcal{L}_{cls}(p_j, \ell_j) +  \lambda {1\over N_{reg}}\sum_j 1_{\ell_j\geq 1}\mathcal{L}_{reg}(t_j, t_j^*)
\end{align}
where $j$ is the index of an anchor (for training the RPN with the ground-truth label $\ell_j\in\{0, 1\}$) or an object-RoI (for training the R-CNN with $\ell_j\in [0, L]$) in a mini-batch, $p_j$ the predicted probability (i.e., soft-max score) of the anchor or the object-RoI being a category $\ell_j$ and $\mathcal{L}_{cls}(p_j, \ell_j)=-\log p_j$, $t_j$ and $t_j^*$ the predicted  $4$-$d$ bounding box regression vector and and ground-truth one respectively and $\mathcal{L}_{reg}(t_j, t_j^*)$ uses the smooth-$l_1$ loss proposed in~\cite{fast_rcnn}.  The term $1_{\ell_j\geq 1}$ means that we only take into account the bounding box regression loss of positives. $N_{cls}$ is usually set to the size of mini-batch and $N_{reg}$ the number of anchor positions in training RPN and the number of object-RoIs in training R-CNN. $\lambda$ is a trade-off parameter to balance the two types of losses.

\textit{Implementation Details.} We follow the widely used practice in specifying the network architecture. For example., when ResNet101~\cite{resNet} is used, we use the first $4$ stages as the shared feature backbone for the RPN and the R-CNN head classifier. The RPN is specified exactly as plotted in Fig.~\ref{fig:network}. After RoICtxMining, the R-CNN head-classifier consists of the final stage of ResNet101 and the two fully-connected branches for prediction and bounding box regression (see Fig.~\ref{fig:network}).

\begin{figure} [t]
\centering
%\framebox
{\includegraphics[width = 0.9\textwidth]{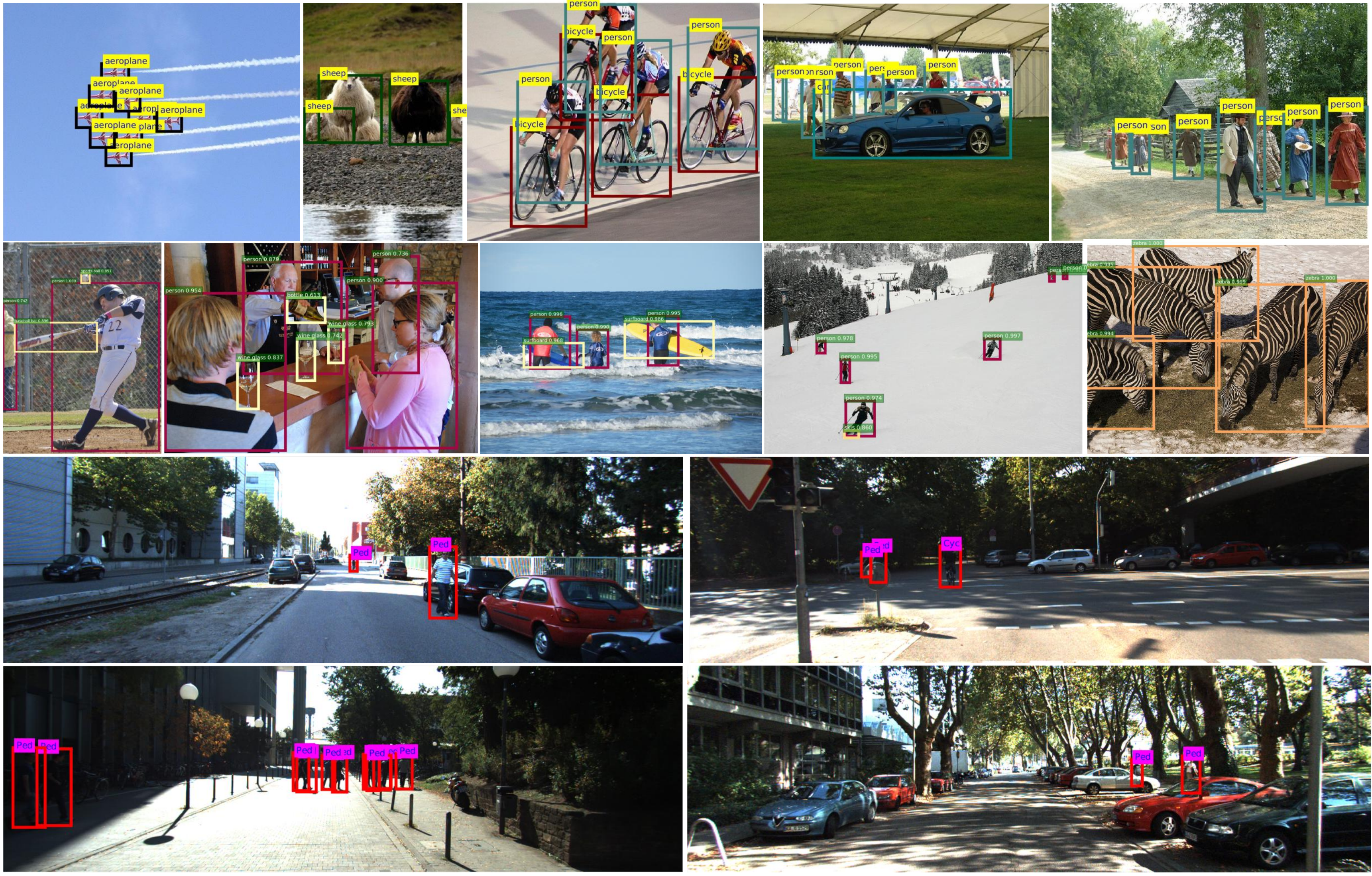}}
\caption{Sample detections of our Auto-Context R-CNN on PASCAL VOC (1st row), MS COCO (2nd row), KITTI pedestrian and cyclist (3rd and 4th rows). Best viewed in color and magnification.}
\label{fig:dets} %\vspace{-4mm}
\end{figure} 

\section{Experiments}
In this section, we first present quantitative and qualitative results of our Auto-Context R-CNN models on PASCAL VOC 2007 and 2012~\cite{pascal}, MS COCO detection benchmark~\cite{coco} and KITTI pedestrian and cyclist detection benchmark~\cite{kitti}. Then, we give detailed ablation studies on the proposed RoICtxPooling operator. 

\textbf{Settings.} We use the ResNet~\cite{resNet} pretrained on the ImageNet~\cite{imagenet} as our feature backbone to fine-tune. We implement our RoICtxPooling operator on top of the RoIPooling operator~\cite{fast_rcnn}, which shows  convincing experimental results for its effectiveness in object detection. Due to the limited GPU computing resource we have, we leave the experiments using the RoIAlign~\cite{mask_rcnn} operator in the future work. We also leave the integration with the deformable variants~\cite{deformable_cnn} of RoIPooling and RoIAlign in the future work. Both are straightforward to implement and should further improve performance.

\begin{table} [t]
\begin{center}
% \resizebox{1.0\hsize}{!}{
\begin{tabular}{|c|c|c|c|c|}
%\hline
%\multicolumn{9}{|c|}{Car-Fluent Recognition} \\
\hline
\multicolumn{4}{|c|}{PASCAL VOC 2007 }  \\
\hline
Method & training data & AP$@0.5$ & AP$@0.7$\\
\hline
% \cite{yuting}  & $07$ & 68.5 & $43.7$  \\
% LocNet \cite{locNet}  & $07$+$12$ & 78.4 & $65.4$  \\
Faster-R-CNN-ResNet101 \cite{faster_rcnn} & $07$+$12$ & $76.4$ & -  \\
R-FCN-ResNet101 & $07$+$12$ & $79.5$ & $60.5$ \\
D-R-FCN-ResNet101$^*$~\cite{deformable_cnn}  & $07$+$12$ & 82.6 & 68.5  \\
Ours-ResNet101 & $07$+$12$ & $\mathbf{83.8}$ & $\mathbf{70.3}$ \\
\hline
R-FCN-ResNet50 \cite{rfcn} & $07$+$12$ & $77.4$ & $57.8$  \\
Ours-ResNet50 & $07$+$12$ & $\underline{\mathit{81.7}}$ & $\underline{\mathit{66.6}}$ \\
\hline
\end{tabular}
% }
\end{center}
\caption{Result comparisons on the PASCAL VOC 2007 test set. $^*$ We note that the ResNet101 backbone used in the deformable variant of R-FCN has deformable convolution layers in the last two stages which are more powerful than the vanilla version of ResNet101 used in our model.}
\label{tab:07} 
\vspace{-5mm}
\end{table}

\subsection{Results on Pascal VOC Datasets}
\textbf{PASCAL VOC 2007 Testset}. We first verify our method on the PASCAL VOC 2007 dataset \cite{pascal}. Following \cite{faster_rcnn,rfcn}, the union set of VOC 2007 \textit{trainval} and VOC 2012 \textit{trainval} (``07+12") are used for training, and the VOC 2007 \textit{test} set is used for testing. 
During testing, non-maximum suppression (NMS) is used to report the final results.
We evaluate our model by the standard mean average precision (mAP) and adopt the PASCAL VOC evaluation protocol \cite{pascal}, i.e., a detection is correct only if the intersection over union (IoU) of its bounding box and the ground-truth bounding box are equal to or greater than $0.5$ or $0.7$ (i.e., AP$@0.5$ and AP$@0.7$). 

In this experiment, our Auto-Context R-CNN is trained using a learning rate of $0.001$ for the first $80k$ iterations and $0.0001$ for another $40k$ iterations with a mini-batch size of $2$.

\begin{figure} [t]
\centering
%\framebox
{\includegraphics[width = 1.0\textwidth]{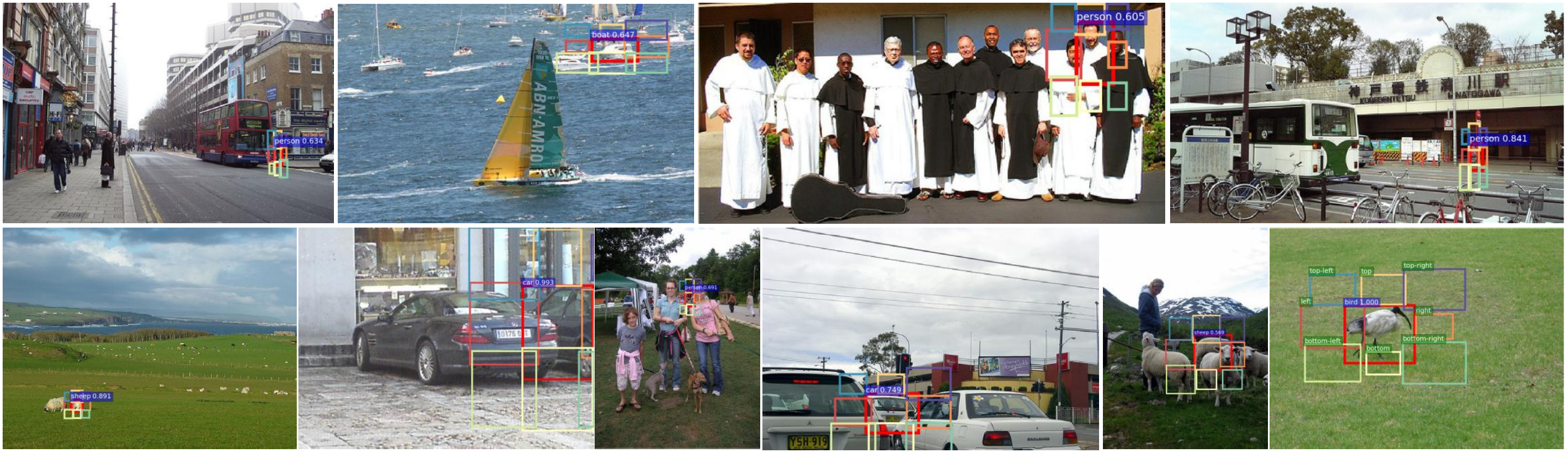}}
\caption{Examples of the proposed context mining method. For clarity, we show only one detection (red) with mined surrounding context RoIs in an image even there are are multiple objects. Best viewed in color and magnification.}
\label{fig:ctx_dets} %\vspace{-3mm}
\end{figure} 

\begin{table} 
\begin{center}
\begin{tabular}{|c|c|c|}
\hline
\multicolumn{3}{|c|}{PASCAL VOC 2012} \\
\hline
Method & training data & AP$@0.5$\\
\hline
YOLOv2 544 ~\cite{yolo}  &$07$++$12$ &  73.4  \\
SSD512~\cite{ssd} &$07$++$12$ &  74.9  \\
Faster-R-CNN-ResNet101 &$07$++$12$ &  $73.8$  \\
R-FCN-ResNet101-ReIm~\cite{rfcn}$^\dag:$ & $07$++$12$ & $76.7$  \\
Ours-ResNet101$^\ddag:$ & $07$++$12$ & $\mathbf{81.1}$  \\
\hline
\end{tabular}
\end{center}
\caption{Result comparisons on the PASCAL VOC 2007 test set. $^\dag$: we use the author provided code to test (\url{https://github.com/daijifeng001/R-FCN}). $^\ddag$: the results of our model can be viewed at \small{\url{http://host.robots.ox.ac.uk:8080/anonymous/PVPYUX.html}}}
\label{tab:12} \vspace{-4mm}
\end{table}

Table \ref{tab:07} shows the comparisons between our model and three state-of-the-art baseline models, vanilla Faster R-CNN~\cite{fast_rcnn}, the fully convolutional variant, R-FCN~\cite{rfcn} and its latest deformable extension~\cite{deformable_cnn}.
Our model outperforms all of them. Compared with the vanilla Faster R-CNN~\cite{faster_rcnn}, we only substitute the RoIPooling operator with our RoICtxMining one. We obtain significant improvement by $7.4\%$ AP$@0.5$ which justifies the effectiveness of our model. Compared with the R-FCN~\cite{rfcn}, it utilizes position-sensitive maps in RoIPooling. Our model outperforms it when using both the ResNet50 and ResNet101 feature backbones. In terms of AP$@0.5$, we obtain $4.3\%$ improvement for both backbones. In terms of AP$@0.7$, we improve it by $8.8\%$ and $9.8\%$ respectively, which also shows the effectiveness of accurate detection of our model. Compared with the deformable variant of R-FCN, our model improves it by $1.2\%$ and $1.8\%$ respectively. We expect that we will be able to further improve the performance through the integration of our RoICtxMining operator and the deformable convolution (which is out of the scope of this paper).

\textbf{PASCAL VOC 2012 Testset}.
We also evaluate our model on PASCAL VOC 2012 benchmark \cite{pascal},and use VOC 2007 \textit{trainval+test} and VOC 2012 \textit{trainval} (``07++12") following \cite{faster_rcnn,rfcn} as our training set, and test on VOC 2012 \textit{test}. 
Training and testing settings are the same as VOC 2007. 

In experiments, we use ResNet101 as the backbone architecture. Our model is trained using a learning rate of $0.001$ for the first $100k$ iterations and $0.0001$ for another $50k$ iterations with a mini-batch size of $2$. 

Table \ref{tab:12} shows the comparisons between our model and some state-of-the-art methods. Our model outperforms all of them. 
Compared with R-FCN, the improvement of our model on PASCAL VOC 2012 is almost the same as the one on PASCAL VOC 2007 dataset, $4.4\%$ vs. $4.3\%$, which shows the stability of performance improvement of our model.

Fig. \ref{fig:dets} shows some qualitative results of our model on the PASCAL VOC testset (1st row).

\subsection{Results on Microsoft COCO Dataset}
To further validate our model on larger datasets, we test it on the MS COCO dataset~\cite{coco} with ResNet-101 as the backbone. There two settings for this experiment.

\textbf{COCO train-val Dataset.} In the first setting, $80k$ \textit{train} set is utilized for training, and $40k$ \textit{val} set is utilized for testing. The training setting is similar to the one on PASCAL VOC datasets. We use a learning rate $0.002$ for the first $90k$ iterations and $0.0002$ for the next $30k$ iterations, with an effective mini-batch size of $8$. 

\begin{table} [t]
\begin{center}

\begin{tabular}{|c|c|c|c|c|c|c|}
\hline
%\multicolumn{9}{|c|}{Car-Fluent Recognition} \\
Method &   AP$@0.5$ & AP$@[0.5:0.95]$ & Small & Medium & Large \\
\hline
Faster-R-CNN-ResNet101~\cite{faster_rcnn} &  $48.4$ & $27.2$ & 6.6 & 28.6 & 45.0 \\
R-FCN-ResNet101~\cite{rfcn} &  $48.9$ & $27.6$  & 8.9 & 30.5 & 42.0  \\
Ours-ResNet101 &  $\mathbf{54.4}$ & $\mathbf{33.7}$ & $\mathbf{15.5}$ & $\mathbf{36.3}$ & $\mathbf{46.2}$  \\ \hline
\end{tabular}

\end{center}
\caption{Result comparisons on the COCO \textit{val} dataset.}
\label{tab:coco_val} \vspace{-5mm}
\end{table}

\begin{table} 
\begin{center}
\resizebox{1.0\hsize}{!}{
\begin{tabular}{|c|c|c|c|c|c|c|c|}
\hline
%\multicolumn{9}{|c|}{Car-Fluent Recognition} \\
Method &  data & AP$@0.5$ & AP$@0.75$ & AP$@[0.5:0.95]$ & Small & Medium & Large\\
\hline
Fast R-CNN & trainval & 39.9 & 19.4 & 20.5 & 4.1 & 20.0 & 35.8 \\
ION~\cite{ion} & trainval35k & 43.2 & 23.6 & 23.6 & 6.4 & 24.1 & 38.3 \\
YOLOv2~\cite{yolo} & trainval35k & 44.0 & 19.2 & 21.6 & 5.0 & 22.4 & 35.5 \\
SSD512~\cite{ssd}  & trainval35k & $46.5$ & 27.8 & $26.8$ & 9.0 & 28.9 & 41.9  \\
R-FCN-ResNet101~\cite{rfcn} & trainval & $51.5$ & - & $29.2$ & 10.3 & 32.4 & 43.3 \\
Faster-R-CNN-ResNet101~\cite{deformable_cnn} & trainval & - & - & $29.4$ & 9.0 & 30.5 & 47.1 \\
R-FCN-ResNet101~\cite{deformable_cnn} & trainval & - & - & $30.8$ & 11.8 & 33.9 & 44.8 \\
D-Faster-R-CNN-ResNet101~\cite{deformable_cnn} & trainval & - & - & 33.1 & 11.6 & 34.9 & $\mathbf{51.2}$ \\
D-R-FCN-ResNet101~\cite{deformable_cnn} & trainval & - & - & 34.5 & 14.0 & 37.7 & 50.3 \\
Faster-R-CNN-FPN-ResNet101~\cite{FPN} & trainval & $\mathbf{59.1}$ & - & $\mathbf{36.2}$ & $\mathbf{18.2}$ & 39.0 & 48.2 \\
Ours-ResNet101 & trainval35k & $57.2$ & $\mathbf{39.1}$ & $36.1$ & 17.3 & \textbf{39.2} & 49.6 \\ \hline
\end{tabular}
}
\end{center}
\caption{Detection results of baseline models and CM-CNN on COCO \textit{test-dev} dataset.}
\label{tab:coco_test_dev} %\vspace{-6mm}

\end{table}

Table \ref{tab:coco_val} shows the comparisons on COCO \textit{val} set. Our model outperforms Faster-RCNN and R-FCN on both AP$@0.5$ and AP$@[0.5:0.95]$ evaluation protocols, where AP$@[0.5:0.95]$ is a new evaluation metric proposed in the COCO benchmark, that averages mAP over different IoU thresholds,
from $0.5$ to $0.95$, thus places a significantly larger emphasis on localization compared to the single AP$@0.5$ or AP$@0.7$.
Our model improves Faster R-CNN and R-FCN by  $6.0\%$ and $5.5\%$ AP$@0.5$ and $6.5\%$ and $6.1\%$ AP$@[0.5:0.95]$, respectively.
For the small, medium and large sets, our model also outperforms Faster R-CNN and R-FCN. Especially, the improvement on the small set is more significant than the one on the large set (e.g., $8.9\%$ vs. $1.2\%$ for Faster R-CNN, and $6.6\%$ vs. $4.2\%$ for R-FCN), \textit{which shows the importance of surrounding contextual information helping small object detection.} 

\textbf{COCO test-dev Dataset.} In the second setting, we use the \textit{trainval35k} in training as done in~\cite{ion} which use all the $80k$ \textit{train} images and $35k$ \textit{val} images (out of total $40k$), and test our model on \textit{test-dev}. We use a learning rate $0.002$ for the first $120k$ iterations and $0.0002$ for the next $50k$ iterations, with an effective mini-batch size of $8$. 

Table \ref{tab:coco_test_dev} shows the comparison results. Similar to results in COCO train-val setting, our model outperforms Faster R-CNN and R-FCN baselines significantly on high precise localization (above $4.3\%$ on AP$@[0.5:0.95]$) and small object detection\footnote{Small means area $<32^2$ pixels; about $40\%$ of COCO objects are small.} (above $5.5\%$ on AP$^{Small}$). This verify the effects of context modeling in our model.
Our model also outperforms recently proposed deformable Faster R-CNN (D-Faster-R-CNN-ResNet101) and deformable R-FCN (D-R-FCN-ResNet101) \cite{deformable_cnn} by $3.0\%$ and $1.7\%$ on AP@[0.5:0.95] respectively. For small objects, our model outperforms D-Faster-R-CNN-ResNet101 and D-R-FCN-ResNet101 by $5.7\%$ and $3.3\%$  respectively.
Besides, our model is also on-par with state-of-the-art feature pyramid networks (FPN), $36.1$ vs $36.2$ on AP@[0.5:0.95]. We note that the proposed RoICtxMining can be integrated with deformable convolutional networks and feature pyramid networks for the best of all (which is out of the scope of this paper).  
Fig. \ref{fig:dets} shows some qualitative results of our model on the MS COCO testset (2nd row). 

\begin{table} [t]
\begin{center}

\begin{tabular}{|c|c|c|c|c|c|c|}
%\hline
%\multicolumn{9}{|c|}{Car-Fluent Recognition} \\
%\hline
%\multicolumn{3}{|c|}{KITTI 2D Detection Benchmark}  \\
\hline
\multirow{2}{*}{Method} & \multicolumn{3}{c|}{Pedestrians} & \multicolumn{3}{c|}{Cyclists} \\
\cline{2-7}
& Easy & Mod & Hard & Easy & Mod & Hard \\
\hline 
Faster R-CNN~\cite{fast_rcnn} & 78.35 & 65.91 & 61.19 & 71.41 & 62.81 & 55.44 \\
\hline
3DOP~\cite{3DOP} & 82.36 & 67.46 & 64.71 & 80.17 & 68.81 & 61.36 \\
\hline
IVA~\cite{IVA} & 83.03 & 70.63 & 64.68 & 77.63 & 67.36 & 59.62  \\
\hline
GN~\cite{GN} & 80.73 & 71.55 & 64.82 & - & - & -  \\
\hline
SubCNN~\cite{SubCNN} & 83.17 & 71.34 & 66.36 & 77.82 & 70.77 & 62.71 \\
\hline
SDP+RPN~\cite{SDPRPN} & 79.98 & 70.20 & 64.84 & 81.05 & 73.08 & 64.88 \\
\hline
MS-CNN~\cite{MS-CNN} & 83.70 & 73.62 & 68.28 & 82.34 & 74.45 & 64.91  \\
\hline
RRC~\cite{RRC} & 84.14 & 75.33  & 70.39 & 84.96 & 76.47 & 65.46 \\
\hline
Ours & \textbf{87.69} & \textbf{79.75} & \textbf{74.56} & $\mathbf{86.06}$ & $\mathbf{78.21}$ & $\mathbf{69.47}$ \\
\hline
\end{tabular}

\end{center}
\caption{Results on the KITTI benchmark test set (only published works shown).}
\label{tab:kitti} 
\vspace{-4mm}
\end{table}

\subsection{Results on KITTI Pedestrian and Cyclist Detection Benchmark}
To elaborate the importance of context mining, we further evaluate our model on KITTI \cite{kitti} benchmark since it  contains many small and overlapped objects (see the 3rd and 4th rows in Fig.~\ref{fig:dets}). There are three object categories, car, pedestrian, and cyclist on KITTI 2D detection task. We test our model on the more challenging pedestrian and cyclist detection (as the mAP of pedestrian and cyclist in much lower than the one of car in the leadboards). 

Table \ref{tab:kitti} shows the comparison results between our model and state-of-the-art published works. \textit{By the time we writing this paper, our model ranks the first place on both pedestrian and cyclist 2D detection challenge among both published and anonymous submissions\footnote{\url{http://www.cvlibs.net/datasets/kitti/eval\_object.php?obj\_benchmark=2d}. }.}  

Compared with Faster R-CNN, our model improves the performance by a even larger margin, $13.84\%$ AP for the Moderate pedestrian set and $15.40\%$ AP on the Moderate cyclist detection. 
Our model also outperforms the state-of-the-art published work, RRC~\cite{RRC}, which models contextual information in a recurrent manner. This shows the effectiveness of the explicit modeling of contextual information in our method. In addition, our model is much faster than RRC. Our model takes $0.6$ seconds on a $2560\times 768$ image, while RRC takes about $3.6$ seconds (subject to some variations of hardware and actual implementation).

\subsection{Ablation Study}
We perform the ablation study on three aspects: How important is the proposed context mining method (i.e., the RoICtxMining operator)? How vulnerable is our Auto-Context R-CNN to different types of attacks? 
%And, how helpful is the extra-rounds of bounding box regression in training?

\begin{table} [t]
\begin{center}
\resizebox{1.0\hsize}{!}{
\begin{tabular}{|c|c|c|c|c|c|c|c|c|}
%\hline
%\multicolumn{9}{|c|}{Car-Fluent Recognition} \\
%\hline
%\multicolumn{3}{|c|}{KITTI 2D Detection Benchmark}  \\
\hline
Method & w/O context & Local & Local+Global & 4-Neighbor & 8-Neighbor & RoICtxMining & RoICtxMining + Local + Global \\
\hline
AP$@0.5$ & 78.0 & 78.4 & 78.5 & 78.8 & 79.0 & \textbf{81.6} & 81.2 \\
\hline
\end{tabular}
}
\end{center}
\caption{Effects of different choices of context modeling in R-CNNs.}
\label{tab:context} \vspace{-6mm}
\end{table}

\begin{table} 
\begin{center}

\begin{tabular}{|c|c|c|c|c|c|c|}
%\hline
%\multicolumn{9}{|c|}{Car-Fluent Recognition} \\
%\hline
%\multicolumn{3}{|c|}{KITTI 2D Detection Benchmark}  \\
\hline
Method & Original & Black & Random & Flip & Adversarial Patch~\cite{adversarialPatch} \\
\hline
R-FCN & 77.4 & 35.0 & 42.1 & 64.1 & 57.4 \\
\hline
Ours & \textbf{81.7} & \textbf{51.5} & \textbf{52.4} & \textbf{70.5} & \textbf{65.9} \\
\hline
\end{tabular}

\end{center}
\caption{Effects of Auto-Context on Attacking Examples.}
\label{tab:attack} \vspace{-6mm}
\end{table}

\begin{table} 
\begin{center}
\begin{tabular}{|c|c|c|c|c|c|c|}
%\hline
%\multicolumn{9}{|c|}{Car-Fluent Recognition} \\
%\hline
%\multicolumn{3}{|c|}{KITTI 2D Detection Benchmark}  \\
\hline
Method & Original & Black & Random & Flip & Adversarial Patch\\
\hline
w/o context & 78.0 & 39.3 & 45.0 & 65.8 & 57.8 \\
\hline
local context & 78.4 & 37.7 & 42.8 & 65.0 & 56.1 \\ \hline
local+global context & 78.5 & 36.5 & 43.6 & 65.4 & 53.5 \\ \hline
8-neighbor context & 79.0 & 38.1 & 45.0 & 65.7 & 57.2 \\ \hline 
Ours & \textbf{81.7} & \textbf{51.5} & \textbf{52.4} & \textbf{70.5} & \textbf{65.9} \\
\hline
\end{tabular}
\end{center}
\caption{Effects of different context methods on Attacking Examples.}
\label{tab:attack-differentCtx} \vspace{-5mm}
\end{table}

\setlength{\textfloatsep}{12pt}
\begin{figure} [t]
\centering
%\framebox
{\includegraphics[width = 0.9\textwidth]{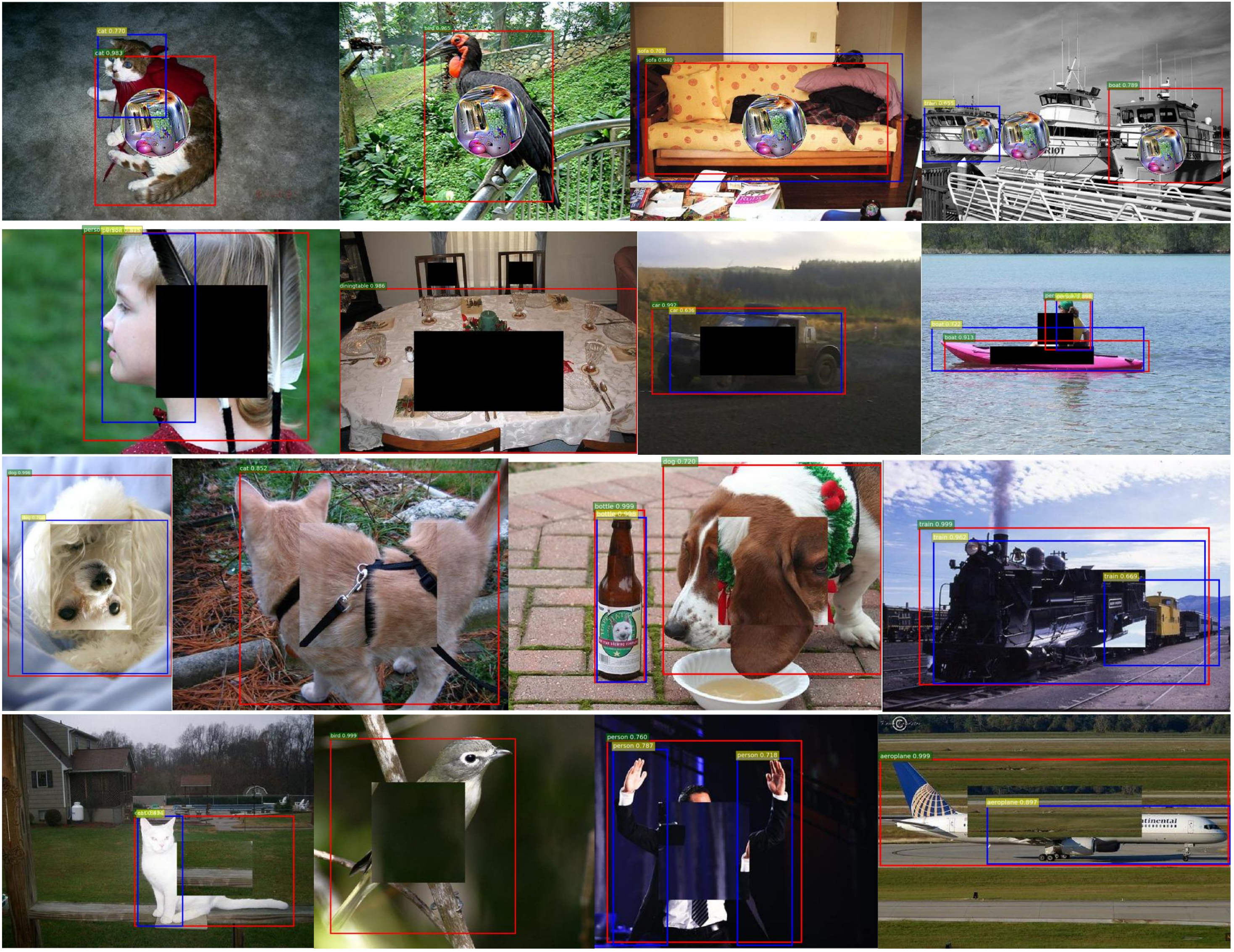}}
\caption{Examples of the four versions of PASCAL VOC 2007 test data and of the detection results by our model (red) and the R-FCN~\cite{rfcn} (blue). The four rows show examples with adversarial-patch~\cite{adversarialPatch}, black-patch, flip-patch and random-patch respectively. Our model shows promising vulnerability to different types of attacks, better than the R-FCN method. Best viewed in color and magnification.}
\label{fig:attack_dets} %\vspace{-6mm}
\end{figure} 

\textit{Effects of Context Mining}. We utilize a $3\times 3$ grid layout when mining contextual information in the $8$ surrounding neighbor cells. To evaluate this design, we compare with other $7$ different choices: \textit{Without context} (i.e., the Faster R-CNN model), \textit{Local context} which utilizes a predefined RoI enlarged from an object-RoI (e.g., $1.5x$ bigger), \textit{Global context} which utilizes the whole feature map as the context RoI, \textit{Local+Global context}, \textit{4-neighbor surrounding context} which utilizes the left, right, top and right cell as context-RoIs in the same $3\times 3$ grid layout, \textit{8-neighbor surrounding context} which utilizes all the $8$ surrounding cells as context-RoIs, and \textit{Local+Global+RoICtxMining}.  We test them on PASCAL VOC 2007 and the proposed context mining method outperforms all other design choices. Predefined local and global context obtain very marginal improvement where the former largely overlaps with the object-RoI and the latter might bring in too much variations. Surrounding contextual information seem help better with similar improvement by 4- or 8-neighbor layouts. Adding the mining component in surrounding contextual regions significantly improves the performance. Surprisingly, adding predefined local+global context to our context mining makes things worse which we will investigate further in our on-going work. Overall, this shows the effectiveness of the current design of our Auto-Context R-CNN. 

Fig.~\ref{fig:ctx_dets} shows sampled results of our model, from which we can see, our model successfully detects small (e.g., the last figure in the first row) and occluded (e.g., the third figure in the first row) objects, which benefiting by the mined contexts which gather supportive information surround the target objects.

\textit{Effects of Auto-Context on Defending Different Attacks}.
Adversarial attacks have become a critical issue for deep learning based approaches including R-CNN based object detection~\cite{attackRCNN,AttackDetectors}. As contextual information is complementary to information inside object RoIs, we expect that our model is more robust to objects attacked by different destructive methods. 
To this end, we first create four versions of PASCAL VOC 2007 test datasets by adding four types of patches to the center of each object instance. The patch is of half width and half height of the annotated bounding box. The four types of patches are: i) \textit{Adversarial-Patch} generated by~\cite{adversarialPatch}, ii) \textit{Black-Patch} with all pixels being zero, iii) \textit{Flip-Patch} that is a flipped version of the center patch itself using left/right, top/bottom, or both randomly, and iv) \textit{Random-Patch} that randomly sampled from outside of a bounding box.

We compare our model with the R-FCN~\cite{rfcn}, both using ResNet50 as feature backbone (see Tabel~\ref{tab:07}). 
Table \ref{tab:attack} shows the comparison results. Although both models are significantly affected by the attacks, our model is more robust than the R-FCN on all the $4$ cases. Fig.~\ref{fig:attack_dets} shows some examples. We also compare different types of context modeling method in Table~\ref{tab:attack-differentCtx}. The results consistently support that the proposed method performs better in adversarial defense due to its auto-context mining. Note that all the models are not adversarially trained.

\section{Conclusion}
This paper presented the Auto-Context R-CNN object detection system which extends the widely used Faster R-CNN~\cite{faster_rcnn} with a novel context mining RoI opeartor (i.e., RoICtxMining). The RoICtxMining operator is a simple yet effective two-layer extension of the popular RoIPooling~\cite{fast_rcnn} or RoIAlign~\cite{mask_rcnn} operator. It is motivated by the Auto-Context work~\cite{auto_context} and the multi-class object layout work~\cite{nms_context}. It explicitly mines surrounding context-RoIs in a $3\times 3$ layout centered at an object-RoI on-the-fly, thus provides an adaptive way of capturing surrounding context.  In experiments, we test our Auto-Context R-CNN using RoIPooling as the backbone RoI operator. It shows competitive results on Pascal VOC 2007 and 2012~\cite{pascal}, Microsoft COCO~\cite{coco}, and KITTI datasets~\cite{kitti} (including $6.9\%$ mAP improvements over a comparative R-FCN~\cite{rfcn} baseline on COCO \textit{test-dev} dataset and the first place on both KITTI pedestrian and cyclist detection as of this submission).

\bibliographystyle{splncs}
\bibliography{arc_fcn}

\begin{thebibliography}{10}

\bibitem{fast_rcnn}
Girshick, R.:
\newblock Fast r-cnn.
\newblock In: International Conference on Computer Vision ({ICCV}). (2015)

\bibitem{faster_rcnn}
Ren, S., He, K., Girshick, R., Sun, J.:
\newblock {Faster R-CNN: Towards Real-Time Object Detection with Region
  Proposal Networks}.
\newblock In: NIPS. (2015)

\bibitem{mask_rcnn}
He, K., Gkioxari, G., Doll{\'{a}}r, P., Girshick, R.B.:
\newblock Mask {R-CNN}.
\newblock In: {IEEE} International Conference on Computer Vision, {ICCV} 2017,
  Venice, Italy, October 22-29, 2017. (2017)  2980--2988

\bibitem{deformable_cnn}
Dai, J., Qi, H., Xiong, Y., Li, Y., Zhang, G., Hu, H., Wei, Y.:
\newblock Deformable convolutional networks.
\newblock In: {IEEE} International Conference on Computer Vision, {ICCV} 2017,
  Venice, Italy, October 22-29, 2017. (2017)  764--773

\bibitem{auto_context}
Tu, Z., Bai, X.:
\newblock Auto-context and its application to high-level vision tasks and 3d
  brain image segmentation.
\newblock {IEEE} Trans. Pattern Anal. Mach. Intell. \textbf{32}(10) (2010)
  1744--1757

\bibitem{nms_context}
Desai, C., Ramanan, D., Fowlkes, C.C.:
\newblock Discriminative models for multi-class object layout.
\newblock International Journal of Computer Vision \textbf{95}(1) (2011)  1--12

\bibitem{rfcn}
Dai, J., Li, Y., He, K., Sun, J.:
\newblock {R-FCN}: Object detection via region-based fully convolutional
  networks.
\newblock arXiv preprint arXiv:1605.06409 (2016)

\bibitem{SS}
Uijlings, J.R.R., van~de Sande, K.E.A., Gevers, T., Smeulders, A.W.M.:
\newblock Selective search for object recognition.
\newblock International Journal of Computer Vision \textbf{104}(2) (2013)
  154--171

\bibitem{edge_boxes}
Zitnick, C.L., Doll\'ar, P.:
\newblock Edge boxes: Locating object proposals from edges.
\newblock In: ECCV. (2014)

\bibitem{BING}
Cheng, M.M., Zhang, Z., Lin, W.Y., Torr, P.H.S.:
\newblock {BING}: Binarized normed gradients for objectness estimation at
  300fps.
\newblock In: IEEE CVPR. (2014)

\bibitem{attackRCNN}
Xie, C., Wang, J., Zhang, Z., Zhou, Y., Xie, L., Yuille, A.L.:
\newblock Adversarial examples for semantic segmentation and object detection.
\newblock In: {IEEE} International Conference on Computer Vision, {ICCV} 2017,
  Venice, Italy, October 22-29, 2017, {IEEE} Computer Society (2017)
  1378--1387

\bibitem{AttackDetectors}
Lu, J., Sibai, H., Fabry, E.:
\newblock Adversarial examples that fool detectors

\bibitem{ctxPriming}
Torralba, A.:
\newblock Contextual priming for object detection.
\newblock International Journal of Computer Vision \textbf{53}(2) (2003)
  169--191

\bibitem{ctxEmpirial}
Divvala, S.K., Hoiem, D., Hays, J., Efros, A.A., Hebert, M.:
\newblock An empirical study of context in object detection.
\newblock In: 2009 {IEEE} Computer Society Conference on Computer Vision and
  Pattern Recognition {(CVPR} 2009), 20-25 June 2009, Miami, Florida, {USA},
  {IEEE} Computer Society (2009)  1271--1278

\bibitem{ctxTinyImages}
Parikh, D., Zitnick, C.L., Chen, T.:
\newblock Exploring tiny images: The roles of appearance and contextual
  information for machine and human object recognition.
\newblock {IEEE} Trans. Pattern Anal. Mach. Intell. \textbf{34}(10) (2012)
  1978--1991

\bibitem{ThreeChannel-IJCV}
Wu, T., Zhu, S.C.:
\newblock A numerical study of the bottom-up and top-down inference processes
  in and-or graphs.
\newblock International Journal of Computer Vision ({IJCV}) \textbf{93}(2)
  (2011)  226--252

\bibitem{resNet}
He, K., Zhang, X., Ren, S., Sun, J.:
\newblock Deep residual learning for image recognition.
\newblock arXiv preprint arXiv:1512.03385 (2015)

\bibitem{ion}
Bell, S., Zitnick, C.L., Bala, K., Girshick, R.B.:
\newblock Inside-outside net: Detecting objects in context with skip pooling
  and recurrent neural networks.
\newblock In: CVPR. (2016)

\bibitem{mrcnn}
Gidaris, S., Komodakis, N.:
\newblock Object detection via a multi-region {\&} semantic segmentation-aware
  {CNN} model.
\newblock CoRR \textbf{abs/1505.01749} (2015)

\bibitem{DeepContext}
Doersch, C., Gupta, A., Efros, A.A.:
\newblock Unsupervised visual representation learning by context prediction.
\newblock In: 2015 {IEEE} International Conference on Computer Vision, {ICCV}
  2015, Santiago, Chile, December 7-13, 2015, {IEEE} Computer Society (2015)
  1422--1430

\bibitem{DPM}
Felzenszwalb, P., Girshick, R., McAllester, D., Ramanan, D.:
\newblock Object detection with discriminatively trained part-based models.
\newblock PAMI (2010)

\bibitem{pascal}
Everingham, M., Van~Gool, L., Williams, C., Winn, J., Zisserman, A.:
\newblock The pascal visual object classes (voc) challenge.
\newblock IJCV (2010)

\bibitem{coco}
Lin, T., Maire, M., Belongie, S.J., Bourdev, L.D., Girshick, R.B., Hays, J.,
  Perona, P., Ramanan, D., Doll{\'{a}}r, P., Zitnick, C.L.:
\newblock Microsoft {COCO:} common objects in context.
\newblock CoRR \textbf{abs/1405.0312} (2014)

\bibitem{kitti}
Geiger, A., Lenz, P., Urtasun, R.:
\newblock Are we ready for autonomous driving? the kitti vision benchmark
  suite.
\newblock In: CVPR. (2012)

\bibitem{rcnn}
Girshick, R., Donahue, J., Darrell, T., Malik, J.:
\newblock Rich feature hierarchies for accurate object detection and semantic
  segmentation.
\newblock In: CVPR. (2014)

\bibitem{imagenet}
Russakovsky, O., Deng, J., Su, H., Krause, J., Satheesh, S., Ma, S., Huang, Z.,
  Karpathy, A., Khosla, A., Bernstein, M., Berg, A.C., Fei-Fei, L.:
\newblock {ImageNet Large Scale Visual Recognition Challenge}.
\newblock International Journal of Computer Vision (IJCV) \textbf{115}(3)
  (2015)  211--252

\bibitem{sppnet}
Kaiming, H., Xiangyu, Z., Shaoqing, R., Sun, J.:
\newblock Spatial pyramid pooling in deep convolutional networks for visual
  recognition.
\newblock In: European Conference on Computer Vision. (2014)

\bibitem{ohem}
Shrivastava, A., Gupta, A., Girshick, R.:
\newblock Training region-based object detectors with online hard example
  mining.
\newblock In: Conference on Computer Vision and Pattern Recognition ({CVPR}).
  (2016)

\bibitem{MS-CNN}
Cai, Z., Fan, Q., Feris, R.S., Vasconcelos, N.:
\newblock A unified multi-scale deep convolutional neural network for fast
  object detection.
\newblock In Leibe, B., Matas, J., Sebe, N., Welling, M., eds.: Computer Vision
  - {ECCV} 2016 - 14th European Conference, Amsterdam, The Netherlands, October
  11-14, 2016, Proceedings, Part {IV}. Volume 9908 of Lecture Notes in Computer
  Science., Springer (2016)  354--370

\bibitem{FPN}
Lin, T., Doll{\'{a}}r, P., Girshick, R.B., He, K., Hariharan, B., Belongie,
  S.J.:
\newblock Feature pyramid networks for object detection.
\newblock In: 2017 {IEEE} Conference on Computer Vision and Pattern
  Recognition, {CVPR} 2017, Honolulu, HI, USA, July 21-26, 2017, {IEEE}
  Computer Society (2017)  936--944

\bibitem{megdet}
Peng, C., Xiao, T., Li, Z., Jiang, Y., Zhang, X., Jia, K., Yu, G., Sun, J.:
\newblock Megdet: {A} large mini-batch object detector.
\newblock CoRR \textbf{abs/1711.07240} (2017)

\bibitem{carAOG}
Li, B., Wu, T., Zhu, S.C.:
\newblock Integrating context and occlusion for car detection by hierarchical
  and-or model.
\newblock In: ECCV. (2014)

\bibitem{guangchen_cvpr}
Guang~Chen, Yuanyuan~Ding, J.X., Han, T.X.:
\newblock Detection evolution with multi-order contextual co-occurrence.
\newblock In: CVPR. (2013)

\bibitem{hoiem06}
Hoiem, D., Efros, A.A., Hebert, M.:
\newblock {Putting Objects in Perspective}.
\newblock In: IEEE CVPR. (2006)

\bibitem{torralba}
Torralba, A.:
\newblock Contextual priming for object detection.
\newblock IJCV \textbf{53} (2003)  2003

\bibitem{laptev15}
Vu, T., Osokin, A., Laptev, I.:
\newblock Context-aware {CNNs} for person head detection.
\newblock In: International Conference on Computer Vision (ICCV). (2015)

\bibitem{sermanet}
Sermanet, P., Kavukcuoglu, K., Chintala, S., Lecun, Y.:
\newblock Pedestrian detection with unsupervised multi-stage feature learning.
\newblock In: Proceedings of the 2013 IEEE Conference on Computer Vision and
  Pattern Recognition. (2013)

\bibitem{scalable}
Szegedy, C., Reed, S., Erhan, D., Anguelov, D.:
\newblock Scalable, high-quality object detection.
\newblock Technical report, arXiv (2015)

\bibitem{deepid}
Ouyang, W., Wang, X., Zeng, X., Qiu, S., Luo, P., Tian, Y., Li, H., Yang, S.,
  Wang, Z., Loy, C.C., Tang, X.:
\newblock Deepid-net: Deformable deep convolutional neural networks for object
  detection.
\newblock In: 2015 IEEE Conference on Computer Vision and Pattern Recognition
  (CVPR). (2015)

\bibitem{multipath}
Zagoruyko, S., Lerer, A., Lin, T.Y., Pinheiro, P.O., Gross, S., Chintala, S.,
  Dollár, P.:
\newblock A multipath network for object detection.
\newblock In: British Machine Vision Conference (BMVC). (2016)

\bibitem{gbdnet}
Zeng, X., Ouyang, W., Yan, J., Li, H., Xiao, T., Wang, K., Liu, Y., Zhou, Y.,
  Yang, B., Wang, Z., Zhou, H., Wang, X.:
\newblock Crafting gbd-net for object detection.
\newblock CoRR \textbf{abs/1610.02579} (2016)

\bibitem{non_local}
Wang, X., Girshick, R.B., Gupta, A., He, K.:
\newblock Non-local neural networks.
\newblock CoRR \textbf{abs/1711.07971} (2017)

\bibitem{auto_net}
Salehi, S.S.M., Erdogmus, D., Gholipour, A.:
\newblock Auto-context convolutional neural network (auto-net) for brain
  extraction in magnetic resonance imaging.
\newblock IEEE Transactions on Medical Imaging (2017)

\bibitem{RRC}
Ren, J.S.J., Chen, X., Liu, J., Sun, W., Pang, J., Yan, Q., Tai, Y., Xu, L.:
\newblock Accurate single stage detector using recurrent rolling convolution.
\newblock In: 2017 {IEEE} Conference on Computer Vision and Pattern
  Recognition, {CVPR} 2017, Honolulu, HI, USA, July 21-26, 2017, {IEEE}
  Computer Society (2017)  752--760

\bibitem{yolo}
Redmon, J., Divvala, S.K., Girshick, R.B., Farhadi, A.:
\newblock You only look once: Unified, real-time object detection.
\newblock In: CVPR. (2016)

\bibitem{ssd}
Liu, W., Anguelov, D., Erhan, D., Szegedy, C., Reed, S., Fu, C.Y., Berg, A.C.:
\newblock {SSD}: Single shot multibox detector.
\newblock arXiv preprint arXiv:1512.02325 (2015)

\bibitem{3DOP}
Chen, X., Kundu, K., Zhu, Y., Berneshawi, A.G., Ma, H., Fidler, S., Urtasun,
  R.:
\newblock 3d object proposals for accurate object class detection.
\newblock In Cortes, C., Lawrence, N.D., Lee, D.D., Sugiyama, M., Garnett, R.,
  eds.: Advances in Neural Information Processing Systems 28: Annual Conference
  on Neural Information Processing Systems 2015, December 7-12, 2015, Montreal,
  Quebec, Canada. (2015)  424--432

\bibitem{IVA}
Zhu, Y., Wang, J., Zhao, C., Guo, H., Lu, H.:
\newblock Scale-adaptive deconvolutional regression network for pedestrian
  detection.
\newblock In Lai, S., Lepetit, V., Nishino, K., Sato, Y., eds.: Computer Vision
  - {ACCV} 2016 - 13th Asian Conference on Computer Vision, Taipei, Taiwan,
  November 20-24, 2016, Revised Selected Papers, Part {II}. Volume 10112 of
  Lecture Notes in Computer Science. (2016)  416--430

\bibitem{GN}
Jung, S., Hong, K.:
\newblock Deep network aided by guiding network for pedestrian detection.
\newblock Pattern Recognition Letters \textbf{90} (2017)  43--49

\bibitem{SubCNN}
Xiang, Y., Choi, W., Lin, Y., Savarese, S.:
\newblock Subcategory-aware convolutional neural networks for object proposals
  and detection.
\newblock In: 2017 {IEEE} Winter Conference on Applications of Computer Vision,
  {WACV} 2017, Santa Rosa, CA, USA, March 24-31, 2017, {IEEE} (2017)  924--933

\bibitem{SDPRPN}
Yang, F., Choi, W., Lin, Y.:
\newblock Exploit all the layers: Fast and accurate {CNN} object detector with
  scale dependent pooling and cascaded rejection classifiers.
\newblock In: 2016 {IEEE} Conference on Computer Vision and Pattern
  Recognition, {CVPR} 2016, Las Vegas, NV, USA, June 27-30, 2016, {IEEE}
  Computer Society (2016)  2129--2137

\bibitem{adversarialPatch}
Brown, T.B., Man{\'{e}}, D., Roy, A., Abadi, M., Gilmer, J.:
\newblock Adversarial patch.
\newblock CoRR \textbf{abs/1712.09665} (2017)

\end{thebibliography}
\end{document}